\documentclass[10pt]{article}
\usepackage{vdn_report}
\usepackage{graphicx}
\usepackage{amsmath,amssymb}
\usepackage{booktabs}
\usepackage{array}
\usepackage{multirow}
\usepackage{url}
\newcommand{\Diag}{\operatorname{Diag}}
\newcommand{\RMSNorm}{\operatorname{RMSNorm}}
\newcommand{\arxivversion}{}

\providecommand{\tightlist}{}
\definecolor{revisionred}{HTML}{000000}
\DeclareRobustCommand{\rev}[1]{{#1}}

\begin{document}
\thispagestyle{empty}
\enlargethispage{0.4in}

\noindent\begin{minipage}{\textwidth}
\VDNFrontMatter
  {Video DeltaNet: A Video-Native Hybrid Attention for Livestream Video Generation}
  {Haocheng Xi$^{1,*}$, Yiming Xie$^2$, Hexu Zhao$^2$, Yiwen Zhang$^2$, Michael Liu$^{1,2}$, Thomas Creavin$^2$\\
   Kurt Keutzer$^1$, Xiuyu Li$^2$, Zhaoyang Lv$^2$, Chenfeng Xu$^3$, Haiwen Feng$^{1,2}$}
  {$^1$University of California, Berkeley \quad $^2$Impossible, Inc. \quad $^3$University of Texas at Austin}
  {Video diffusion models repeatedly process long spatiotemporal token sequences during denoising, making attention a major computational bottleneck. Linear attention offers an appealing alternative and has been widely adopted in recent large language models, but directly applying it to video models often fails to preserve the fine-grained interactions required for high-quality generation. We present \textbf{Video DeltaNet (VDN)}, which combines local Softmax attention with bidirectional linear memory for long-range video context. Its linear branch introduces \textbf{Video Delta Attention (VDA)}, \rev{which updates memory once per frame by jointly resolving correlated spatial writes, with a non-expansive inherited-state transition for fixed prepared features and gates and without additional frame-size key scaling.} Separate output projections and learnable gates calibrate the two branches, while a staged teacher-alignment recipe progressively introduces the new pathway into pretrained models. We instantiate VDN on MiniMax H3, applying the hybrid to video-to-video interactions while retaining Softmax for interactions involving text or audio. With eight-step distillation and an optimized SGLang serving stack, VDN-H3 completes DiT denoising for a 14.3-second, 768p video in 6.70 seconds on eight NVIDIA B200 GPUs, corresponding to a 14.5$\times$ speedup over the 50-step dense H3 baseline on the same GPU count.}

  \begingroup
    \centering
    \includegraphics[width=\linewidth]{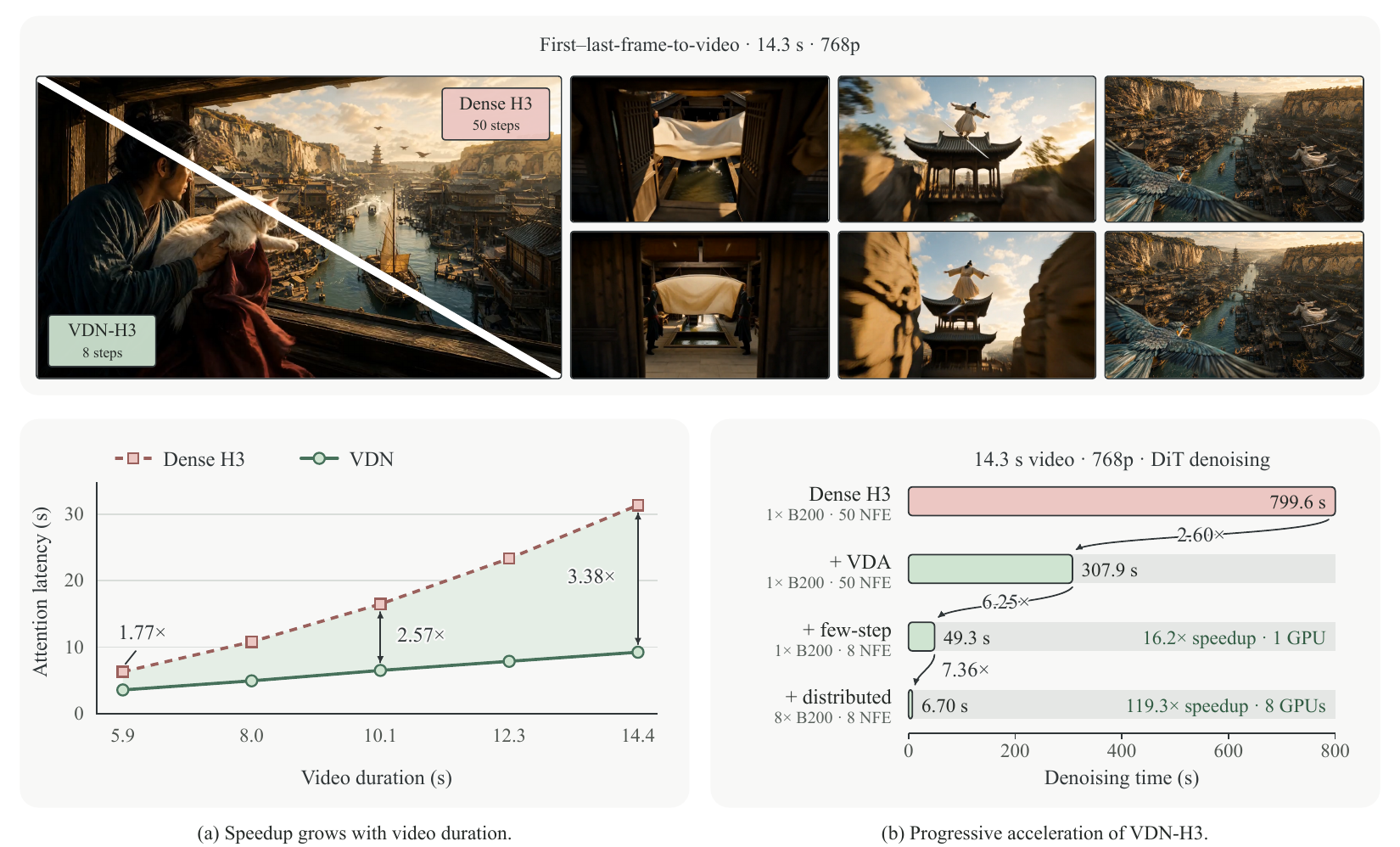}
    \captionof{figure}{Quality and efficiency of VDN-H3. Top: matched video frames. Bottom: attention scaling and cumulative denoising acceleration.}
    \label{fig-teaser}
  \endgroup
\end{minipage}\par

\begingroup
\renewcommand{\thefootnote}{*}
\footnotetext{Part of the work done during an internship at Impossible, Inc.}
\endgroup

\VDNBodyICLR
\section{Introduction}\label{introduction}

Attention is the dominant computational cost in long-sequence video generation. In the profiled MiniMax H3 workload, Softmax attention accounts for more than 85\% of denoiser runtime. Dense Softmax gives every query direct access to the complete key sequence, but the resulting pairwise computation grows quadratically with sequence length. Frontier language models increasingly use recurrent linear attention to avoid this scaling bottleneck. Applying the same idea to video is appealing: distant context can be compressed into a fixed-size state, making its cost linear in the number of tokens. A direct replacement, however, meaningfully degrades generation quality because the compressed state cannot preserve all of the fine-grained interactions available to Softmax.

Closing this quality gap requires addressing three mismatches. First, a fixed-size state must preserve global properties such as subject identity, scene layout, appearance, and long-range motion even though its capacity does not grow with sequence length, whereas Softmax retains an expanding set of keys and values. Second, delta-rule linear attention typically updates its recurrent state one token at a time, mirroring autoregressive language-model decoding. Video diffusion instead processes all spatial tokens in a frame together; imposing an arbitrary patch order is unnatural, while treating correlated writes independently can make them interfere. Third, adding a randomly initialized linear pathway to a Softmax-pretrained model changes both its information flow and residual-stream activation statistics. Without careful adaptation, it can disrupt capabilities learned during pretraining before the new branch becomes useful.

We introduce \textbf{Video DeltaNet (VDN)}, a hybrid attention architecture designed around these challenges. VDN retains Softmax for local video interactions and global boundary anchors, while bidirectional linear memory represents distant video context. The anchors keep the beginning and end of the clip directly accessible as global references. The linear branch uses \textbf{Video Delta Attention (VDA)}, \rev{a video-native delta rule that jointly resolves correlated writes within each frame. For fixed prepared features and gates, its inherited-state transition is non-expansive without scaling keys by the number of spatial tokens per frame.} RMS normalization stabilizes the scale of the linear readout, while separate gates and output projections calibrate each branch before their outputs are combined. A staged adaptation recipe first aligns the new linear pathway with the pretrained teacher, then uses low-rank updates to co-adapt the hybrid layer while preserving the pretrained backbone.

We instantiate the approach on MiniMax H3 to obtain \textbf{VDN-H3}. \ifdefined\arxivversion In collaboration with the SGLang team, we optimize its inference path. \fi With eight-step distillation, VDN-H3 completes DiT denoising for a 14.3-second, 768p video in 6.70 seconds on eight NVIDIA B200 GPUs. This corresponds to a 14.5$\times$ reduction relative to the 50-step dense H3 baseline on the same GPU count.

Our main contributions are:

\begin{enumerate}
\tightlist
\item
  \textbf{A hybrid video attention architecture} that combines local Softmax and global boundary anchors with bidirectional linear memory, together with branch-specific normalization, gates, and output projections.
\item
  \textbf{A frame-wise delta operator} \rev{that jointly resolves correlated spatial writes and yields a non-expansive inherited-state transition for fixed prepared features and gates without additional frame-size key scaling, together with a full derivation and an efficient batched implementation.}
\item
  \textbf{A complete adaptation recipe for pretrained video models} that combines staged teacher alignment, low-rank refinement, few-step distillation, and optimized inference without training a new foundation model from scratch.
\end{enumerate}

\protect\phantomsection\label{architecture}
\section{Video DeltaNet: Hybrid Attention Architecture}\label{video-deltanet-hybrid-attention-architecture}

Video DeltaNet divides video-to-video attention by temporal role. Nearby frames retain explicit token-to-token Softmax attention, while distant video context is summarized by linear attention. Below, we introduce the two branches and explain how their outputs are combined.

\subsection{Sliding-Window Softmax Attention}\label{sliding-window-softmax-attention}

Nearby frames contain local correspondences that determine texture, object boundaries, and short-term motion. These interactions benefit from direct token-to-token matching across neighboring frames. VDN therefore retains exact Softmax attention within a bidirectional temporal window to preserve quality, while assigning distant interactions to linear attention.

The window follows the video tokenizer. H3's VAE decodes five consecutive latent frames as one temporal chunk, so each query chunk attends to itself and its immediately preceding and following chunks. This prevents the Softmax boundary from cutting through the tokenizer's natural temporal unit, producing a 15-frame window except at sequence boundaries.

VDN also adds \textbf{two boundary anchors with four-way connectivity}. Every video frame attends to all tokens in the first and last latent frames, and the first and last frames attend to the complete sequence. This pattern is particularly natural for full-clip video diffusion: the two boundary anchors provide explicit global references from opposite ends of the clip, while only two frame rows and columns receive dense connectivity. The same anchors are valuable for image-to-video and first--last-frame-to-video generation, where the provided visual conditions directly constrain the generated sequence. Anchor entries already inside a local window are included only once.

\begin{figure}[t]
  \centering
  \includegraphics[width=\linewidth]{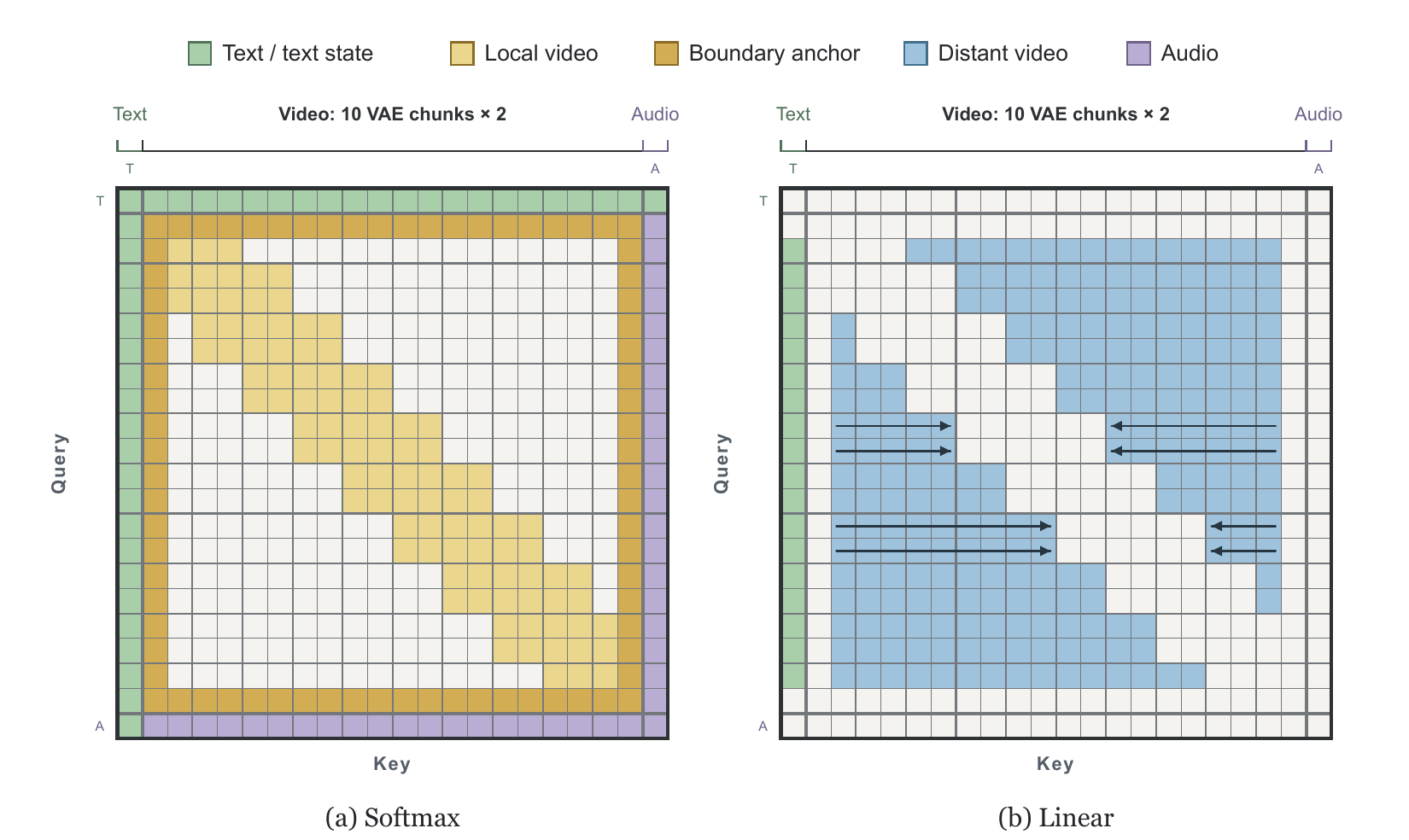}
  \caption{Softmax and Linear attention allocation. The schematic uses one text token, twenty video tokens in ten two-frame chunks, and one audio token.}
  \label{fig-temporal}
\end{figure}

\subsection{Bidirectional Linear Attention}\label{bidirectional-linear-attention}

Bidirectional linear attention handles the remaining video context through two temporal scans. For a query frame $t$, the forward state $S_t^{\rightarrow}$ summarizes frames before its Softmax window, and the reverse state $S_t^{\leftarrow}$ summarizes frames after the window. Boundary anchors are excluded from both states because they are already available through Softmax. The two temporal regions are disjoint, so their query readouts can be added without counting any video frame twice.

Each scan first builds frame states along its own direction. At readout time, VDN gathers the state immediately outside the corresponding window boundary, applies the accumulated channel-wise decay across the skipped local span, and evaluates the query against the resulting memory. The distant-context output is the sum of the two readouts:

\stepcounter{equation}\begin{equation}\label{eq-bidirectional-readout}\tag{1a}
o_t^{L}=S_t^{\rightarrow}q_t+S_t^{\leftarrow}q_t.
\end{equation}

The linear memory is also text-aware. Before scanning the video sequence, VDN summarizes all text tokens into a state $S_T$ and initializes each directional scan with $S_T/2$. Summing the two readouts therefore counts the prompt exactly once, providing the linear branch with global text conditioning while text remains directly visible to the Softmax branch:

\stepcounter{equation}\begin{equation}\label{eq-text-state-init}\tag{1b}
S_0^{\rightarrow}=S_0^{\leftarrow}=\tfrac{1}{2}S_T,
            \qquad
            \big(S_0^{\rightarrow}+S_0^{\leftarrow}\big)q=S_Tq.
\end{equation}

\subsection{Combining Outputs from Two Branches}\label{gates}

Both branches receive their query, key, and value inputs from the pretrained QKV projections. The Softmax branch keeps H3's QK normalization and rotary position processing. Following Gated DeltaNet \citep{yang2024gated} and Kimi Delta Attention \citep{kimiteam2025kimi}, the linear branch then applies its own feature map: a separable short convolution to K and V, followed by SiLU, with L2 normalization on Q and K. The released K/V convolution consists of a depthwise $5\times5$ spatial filter and a five-tap temporal filter. No rotary embedding is added in the linear branch.

The two branches have different output scales and therefore require calibration. Restricting Softmax to local windows and boundary anchors concentrates its probability mass over fewer keys, so VDN applies a content-dependent sigmoid gate to the Softmax readout.

The linear readout is RMS-normalized and passed through its own sigmoid output gate. Following Gated DeltaNet and Kimi Delta Attention, separate decay and write gates control memory retention and frame updates inside the recurrence.

Each branch also has its own output projection. Let the gated branch outputs be

\stepcounter{equation}\begin{equation}\label{eq-branch-gates}\tag{1c}
\widetilde{O}^S=G^S\odot O^S,
      \qquad
      \widetilde{O}^L=G^L\odot\RMSNorm(O^L).
\end{equation}

The two branches are then projected independently and added:

\stepcounter{equation}\begin{equation}\label{eq-fusion}\tag{1d}
Y=\widetilde{O}^S W_O^S+\widetilde{O}^L W_O^L.
\end{equation}

Separate output projections allow the branches to contribute in different residual-stream directions. The combined output is then fed into the pretrained residual stream, while the original feed-forward sublayer is left unchanged.

\begin{figure}[t]
  \centering
  \includegraphics[width=\linewidth]{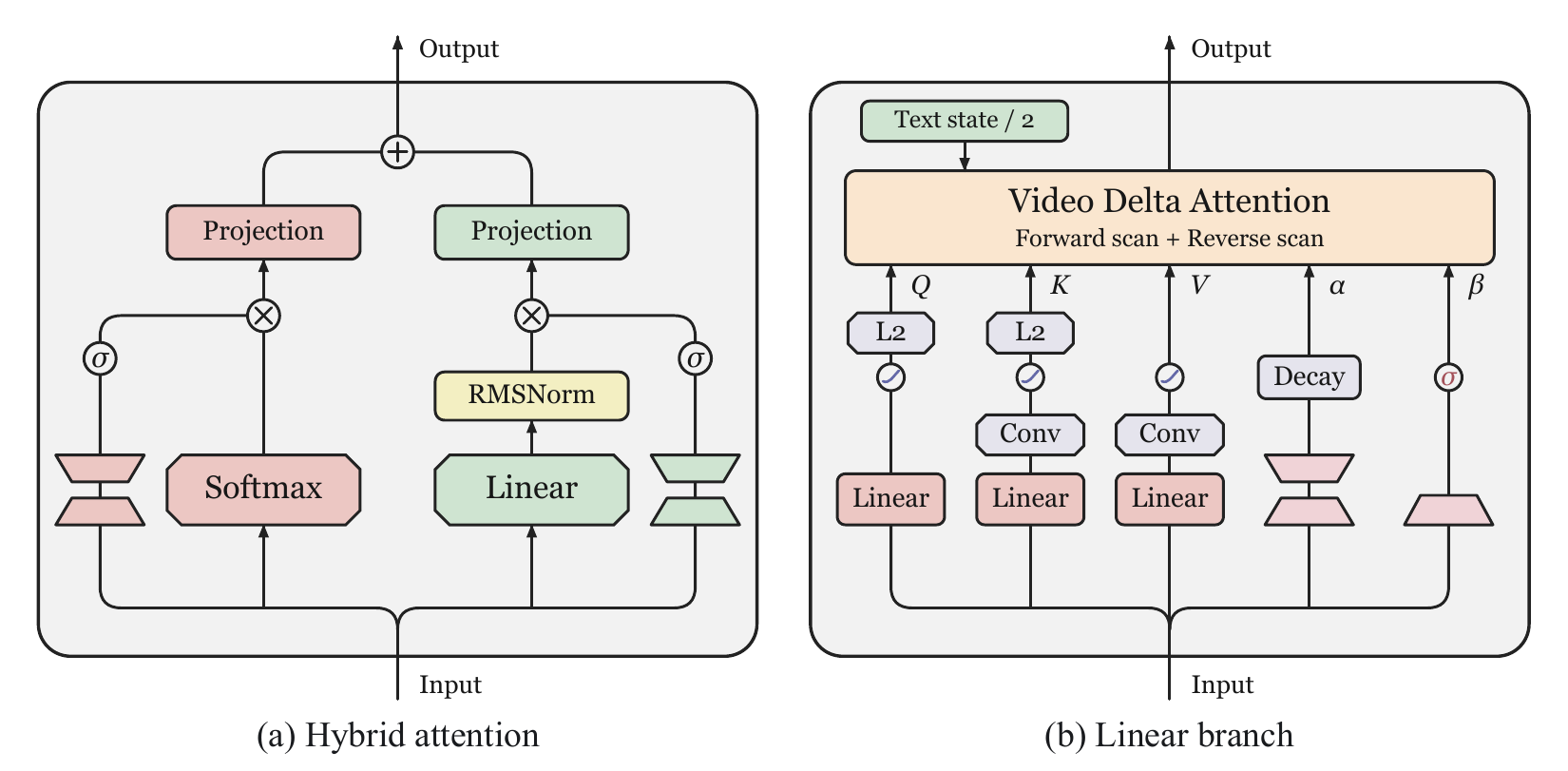}
  \caption{Video DeltaNet architecture. Softmax and Linear outputs are independently gated and projected; the Linear branch uses bidirectional Video Delta Attention.}
  \label{fig-overview}
\end{figure}

\protect\phantomsection\label{vda}
\section{Video Delta Attention: Frame-wise Delta Rule}\label{video-delta-attention-frame-wise-delta-rule}

Video Delta Attention (VDA) extends the delta rule from individual tokens to entire video frames. It jointly updates the recurrent state from all key-value pairs in a frame, allowing interactions among frame tokens to shape the write rather than accumulating independent corrections. We first review the standard token-wise rule and then derive the frame-wise update.

\subsection{Preliminaries of Linear Attention}\label{preliminaries-of-linear-attention}

Delta-rule memory reads a value associated with a key and writes a correction proportional to the prediction residual. Gated DeltaNet combines this mechanism with memory decay \citep{yang2024gated}, while Kimi Delta Attention introduces finer-grained decay \citep{kimiteam2025kimi}. At step $t$, for a state $S_{t-1}\in\mathbb R^{d_v\times d_k}$, a key $k_t\in\mathbb R^{d_k}$, and a value $v_t\in\mathbb R^{d_v}$, the familiar one-token update is

\stepcounter{equation}\begin{equation}\label{eq-delta}\tag{2}
\begin{aligned}
      \bar S_t&=S_{t-1}\Diag(\alpha_t),\\
      S_t&=\bar S_t+\beta_t(v_t-\bar S_t k_t)k_t^\top .
      \end{aligned}
\end{equation}

The decay gate $\alpha_t$ controls how much inherited memory remains, while $\beta_t$ controls the strength of the erase-and-write correction. This token-wise recurrence is well matched to autoregressive decoding, where one new token arrives at each step.

Video diffusion presents a different unit of computation: all spatial tokens of a latent frame are available together. Sequentially imposing a patch order is unnecessary, so a natural first adaptation is to compute their delta corrections in parallel from the same decayed state. SANA-WM follows this batched construction and adds frame-size key scaling to stabilize the resulting additive transition \citep{zhu2026sanawm}.

Let a video contain $F$ latent frames with $U=H_\ell W_\ell$ tokens each. For token $u$ of frame $t$, denote its key, value, and write gate by $k_{t,u}\in\mathbb R^{d_k}$, $v_{t,u}\in\mathbb R^{d_v}$, and $\beta_{t,u}\ge0$. A frozen-state update adds all token corrections evaluated at $\bar S_t=S_{t-1}\Diag(\alpha_t)$:

\stepcounter{equation}\begin{equation}\label{eq-batch-sum}\tag{3a}
S_t^{\mathrm{batch}}=\bar S_t+\sum_{u=1}^{U}\beta_{t,u}(v_{t,u}-\bar S_tk_{t,u})k_{t,u}^{\top}.
\end{equation}

Stack the keys and values as $K_t\in\mathbb R^{U\times d_k}$ and $V_t\in\mathbb R^{U\times d_v}$, and write $\beta_t=(\beta_{t,1},\ldots,\beta_{t,U})^\top$. The two frame statistics are

\stepcounter{equation}\begin{equation}\label{eq-statistics}\tag{3b}
\begin{aligned}
      A_t&=\sum_{u=1}^{U}\beta_{t,u}k_{t,u}k_{t,u}^{\top}=K_t^{\top}\Diag(\beta_t)K_t,\\
      B_t&=\sum_{u=1}^{U}\beta_{t,u}v_{t,u}k_{t,u}^{\top}=V_t^{\top}\Diag(\beta_t)K_t.
      \end{aligned}
\end{equation}

Here $A_t$ summarizes key correlations and $B_t$ value--key writes, giving

\stepcounter{equation}\begin{equation}\label{eq-batch-delta}\tag{3c}
S_t^{\mathrm{batch}}=\bar S_t(I-A_t)+B_t.
\end{equation}

Because every residual in Equation \eqref{eq-batch-sum} uses the same $\bar S_t$, overlapping keys can produce conflicting writes without accounting for one another (Appendix \ref{frame-size-scaling-and-correlation-awareness}).

\subsection{Video Delta Attention: A Frame-wise Update}\label{video-delta-attention-a-frame-wise-update}

This independence becomes problematic when several patches address similar key directions. Their corrections can reinforce or conflict even though they are meant to describe one frame. VDA instead lets all spatial tokens determine a single new state together, so overlapping directions are resolved inside the update while previously accumulated memory remains a reference.

The classical one-token delta rule can be viewed as one gradient step on its prediction error. Rather than taking one such step independently for every patch, VDA defines the frame-level state as the solution to a joint objective:

\stepcounter{equation}\begin{equation}\label{eq-objective}\tag{4}
S_t=\arg\min_S\ \frac12\|S-\bar S_t\|_F^2+\frac12\sum_{u=1}^{U}\beta_{t,u}\|Sk_{t,u}-v_{t,u}\|_2^2.
\end{equation}

The first term keeps the new memory close to the decayed state. The second asks that same state to fit all key--value associations in the frame simultaneously. Differentiating once gives a compact normal equation and closed-form update:

\stepcounter{equation}\begin{equation}\label{eq-solution}\tag{5}
S_t(I+A_t)=\bar S_t+B_t,\qquad S_t=(\bar S_t+B_t)(I+A_t)^{-1}.
\end{equation}

In contrast to Equation \eqref{eq-batch-delta}, each residual in the joint solution is evaluated at the shared updated state $S_t$. The inverse couples the writes: when two patch keys overlap, their inner product affects both effective corrections. We compute this inverse in the $d_k\times d_k$ key-channel space; Appendix \ref{frame-size-scaling-and-correlation-awareness} illustrates the effect of key correlations.

\subsection{Stability and Correlation Awareness of Video Delta Attention}\label{stability}

VDA has a stable inherited-state transition: the contribution carried from earlier frames cannot be amplified by a frame update when the prepared features and gates are fixed.

\textbf{Proposition 1 --- Non-expansive inherited-state transition.} For fixed prepared features and gates, with $\beta_{t,u}\ge0$ and $0\le\alpha_{t,j}\le1$, the transition $M_t=\Diag(\alpha_t)(I+A_t)^{-1}$ satisfies $\|M_t\|_2\le1$. Thus changing the entering state by $\Delta S$ changes its carried contribution by at most $\|\Delta S\|_F$.

\rev{With unit-normalized prepared keys, frozen-state additive frame-wise writes can amplify inherited state. SANA-WM applies an additional $1/\sqrt U$ key scale, replacing the frame statistics $A_t$ and $B_t$ by $A_t/U$ and $B_t/\sqrt U$, respectively \citep{zhu2026sanawm}. VDA instead couples the within-frame writes through $(I+A_t)^{-1}$: for fixed prepared features and gates, this solve makes the inherited-state transition non-expansive without the extra frame-size scale. This is a bound on the carried state contribution, rather than on the full input-dependent network. Appendix \ref{appendix-proof} proves the proposition, and Appendix \ref{frame-size-scaling-and-correlation-awareness} compares the two rules under different key correlations.}

\protect\phantomsection\label{training}
\section{End-to-End Training Pipeline}\label{end-to-end-training-pipeline}

We first describe the three-stage adaptation that integrates the Linear branch into the pretrained Softmax backbone, followed by few-step distillation from 50 to eight denoising steps.

\subsection{Staged Architecture Adaptation}\label{staged-architecture-adaptation}

\begin{figure}[t]
  \centering
  \includegraphics[width=\linewidth]{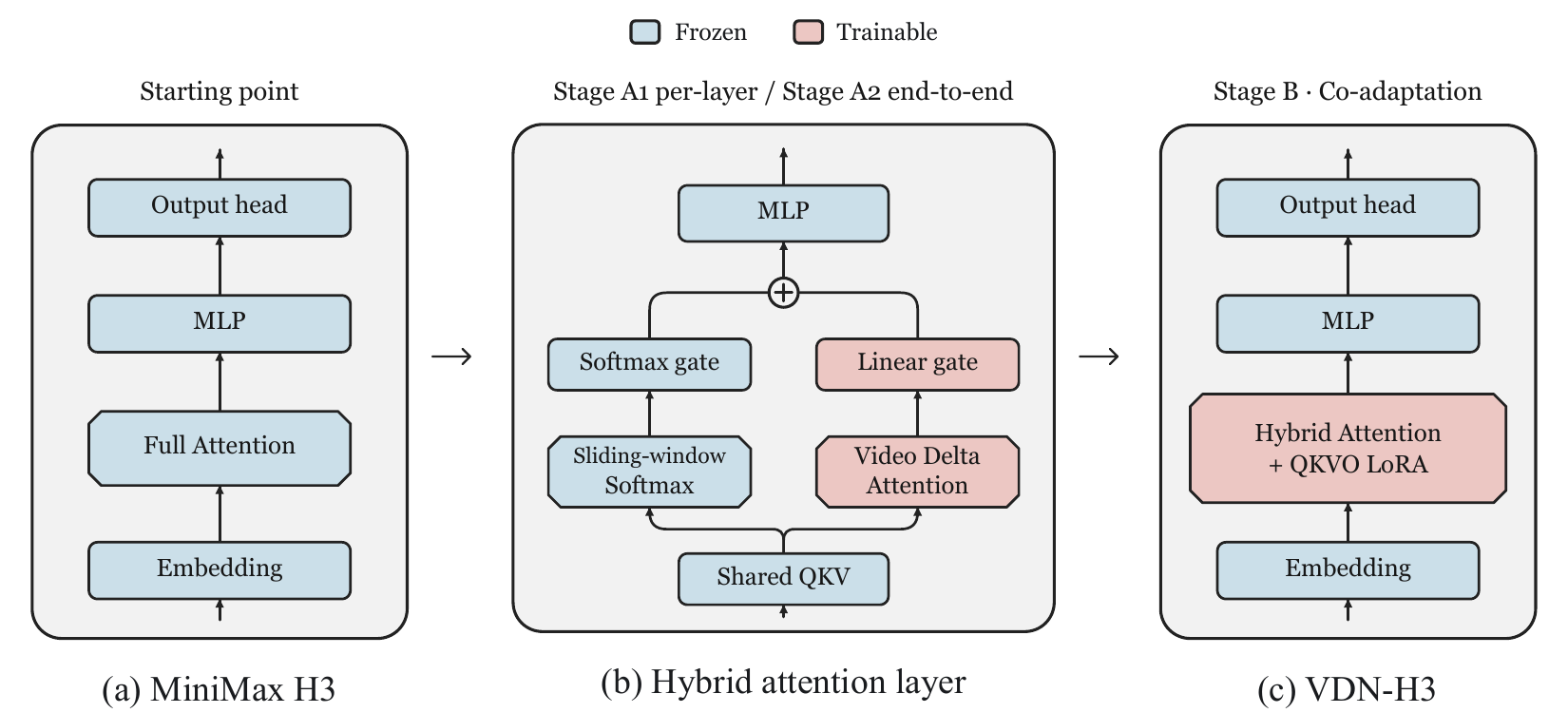}
  \caption{Progressive adaptation of the pretrained H3 backbone. (a) The dense model is the frozen starting point. (b) Stage A1 aligns each Linear branch independently, then Stage A2 trains the assembled hybrid layers end to end while the pretrained weights and Softmax gates remain frozen. (c) Stage B jointly trains the Linear pathway, Softmax gates, and QKVO LoRA adapters.}
  \label{fig-training}
\end{figure}

Architecture adaptation proceeds in three stages. Figure \ref{fig-training} shows the trainable components in each stage; the pretrained base weights remain frozen throughout.

\textbf{A1 - Per-layer alignment.} Each Linear branch is initialized independently from frozen pretrained activations for 200 steps. This local objective avoids sending the initialization signal through a deep stack of simultaneously changing hybrid blocks. The backbone is frozen, the Softmax gate is fixed at its 0.99 initialization, and gradients are clipped independently for each layer at 0.1.

\textbf{A2 - End-to-end alignment.} The calibrated branches are then installed together and optimized end to end for 500 steps. This stage corrects composition errors that are invisible when blocks are trained in isolation. The pretrained weights and Softmax gates remain frozen, and gradient clipping is applied globally at 1.0.

\textbf{Stage B - LoRA co-adaptation.} We add LoRA adapters to the Q, K, V, and output projections and train them jointly with the Linear pathway and Softmax gates for 2,000 steps.

\subsection{Few-Step Distillation}\label{few-step-distillation}

After architecture adaptation, we distill the 50-step VDN-H3 into an eight-step sampler using a DMD2-style objective without the GAN term \citep{yin2024dmd2}. The student is initialized from the community MiniMax-H3-Turbo-LoRA \citep{larryvrh2026turbo} and trained against VDN-H3's own 50-step sampler, isolating step reduction from architecture conversion. Generator, real-score, and fake-score roles share one FSDP backbone through separate adapters, with three fake-score updates per generator update. The released checkpoint is trained for 250 generator steps.

\protect\phantomsection\label{systems}
\section{Efficient Inference}\label{efficient-inference}

\textbf{Fused VDA kernels.} We organize VDA's data preparation and readout into four fused Triton kernels. \emph{VDA-Prep} combines temporal convolution, SiLU, L2 normalization, and the frame-major layout conversion for K and V. \emph{VDA-Stats} constructs the per-frame statistics $A_t$ and $B_t$ in one pass, sharing input reads and reduction work. \emph{VDA-Gather} collects the two directional states at local-window boundaries and applies the decay bridge. \emph{VDA-Epilogue} combines RMS normalization, output gating, and the final layout conversion. QK normalization and rotary embedding are fused separately on the Softmax path.

\textbf{Chunk-wise scans.} VDA defines an affine transition per frame, but attention reads memory only at VAE-chunk boundaries. We therefore compose each chunk into $S_{\mathrm{out}}=S_{\mathrm{in}}M_{\mathrm{chunk}}+J_{\mathrm{chunk}}$ and scan the shorter chunk sequence in both directions with a single kernel launch. This preserves the required boundary states while reducing scan depth and launch overhead by roughly the chunk size. The prompt state is included as a leading virtual frame.

\textbf{Small-matrix inverse.} Each frame requires computing $(I+A_t)^{-1}$. A batched Cholesky pipeline incurs several kernel launches and intermediate memory reads and writes, so we use one CUDA kernel that performs blocked Gauss--Jordan elimination in registers and directly emits the transition and injection terms. The inverse and recurrent state updates remain in FP32.

\textbf{Other optimizations.} \ifdefined\arxivversion The released VDN-H3 model is served through SGLang. \fi Window Softmax packs queries by visible-key pattern to use FlashAttention's varlen API without a global mask. Following Ulysses \citep{jacobs2023ulysses}, VDA is sharded by attention head and executed on a side stream alongside window Softmax. MXFP8 accelerates the wide QKV, output, and feed-forward GEMMs, while recurrent states and small-matrix inverses remain in FP32. AdaLN modulation parameters are precomputed before the block loop to avoid repeating their projection inside each transformer block.

\protect\phantomsection\label{evaluation}
\section{Experiments}\label{experiments}

\subsection{Settings}\label{settings}

\rev{\textbf{Data.} We curated a training set of 10,015 video clips at 1344 {$\times$} 768 resolution, each with 345 frames at 24 fps (14.375 seconds). Each sample is pre-encoded and cached as video latents of shape {$(24,102,48,84)$}, stereo audio latents of shape {$(2,32,575)$}, and Qwen3-VL text embeddings of shape {$(L,5120)$} with token-type tags \citep{bai2025qwen3vl}. This removes the VAEs and text encoder from the training loop. For evaluation, every model renders the same 103 fixed third-party prompts at the same resolution and duration.}

\rev{\textbf{Baselines.} The 50-NFE comparison includes full-attention Dense H3, the training-free sparse baseline Radial Attention \citep{li2025radial}, and the Stage-B VDN-H3 checkpoint. Radial Attention's density is set to match VDN-H3's B200 latency. The 8-NFE comparison includes Dense H3, distilled VDN-H3, and FastH3~v2 \citep{fastvideo2026fasth3v2}, a sparse-attention baseline for video generation. These comparisons isolate architecture changes at 50 NFE and evaluate accelerated samplers at 8 NFE.}

\rev{\textbf{Metrics.} For text-to-video evaluation, we report EvalCrafter VQA\textsubscript{A} and VQA\textsubscript{T} \citep{liu2024evalcrafter}, Q-Align \citep{wu2024qalign}, and DOVER++ overall \citep{wu2023dover}. FIRM-Video evaluates Instruction Following, Perceptual Quality, and World Coherence on a 1--5 scale \citep{zhang2026firmvideo}. For first--last-frame-to-video evaluation (FL2VA), PSNR and SSIM measure fidelity at the two conditioning frames \citep{wang2004ssim}. Higher is better for all quality metrics; Q-Align and DOVER++ use a {$\times100$} scale. We additionally report RAFT mean flow as a motion diagnostic \citep{teed2020raft}.}

\rev{\textbf{Environment.} We use PyTorch 2.13 and CUDA 12.9 on NVIDIA H200 and B200 clusters. Training uses FSDP2/HSDP, activation checkpointing, and pinned-memory activation offload. Parameters are gathered in bf16 and gradients reduced in FP32, except that the decay-{$\alpha$} modules remain FP32. Appendix~\hyperref[appendix-training-hparams]{A.4} gives the architecture-adaptation settings.}

\subsection{Quality Results}\label{quality-results}

\begin{figure}[!htb]
\captionsetup{labelfont={sf,bf,color=revisionred}}
  \centering
  \includegraphics[width=\linewidth]{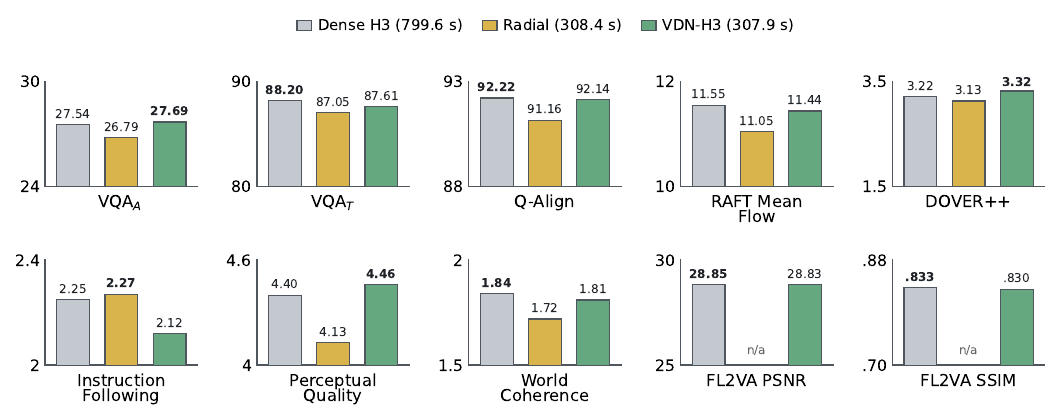}
  \caption{\rev{50-NFE comparison across ten metrics. VDN-H3 remains close to Dense H3 and is higher than Radial on six of seven shared quality metrics while matching its B200 latency. Radial was not evaluated on FL2VA.}}
  \label{fig-quality-50-main}
\end{figure}

\rev{Figure \ref{fig-quality-50-main} compares the attention architectures at 50 NFE across ten metrics. VDN-H3 remains close to Dense H3: it is higher on VQA\textsubscript{A}, DOVER++, and FIRM perceptual quality, while the gaps are only 0.08 on Q-Align, 0.03 on FIRM world coherence, 0.02 dB on FL2VA PSNR, and 0.0024 on FL2VA SSIM. Across the seven quality metrics shared with Radial, VDN-H3 is higher on six, including gains of 0.90 on VQA\textsubscript{A}, 0.98 on Q-Align, and 0.33 on FIRM perceptual quality; RAFT flow remains similar. These results support our claim that VDN-H3 maintains quality comparable to Dense H3 while outperforming Radial at similar latency.}

\begin{figure}[!htb]
\captionsetup{labelfont={sf,bf,color=revisionred}}
  \centering
  \includegraphics[width=\linewidth]{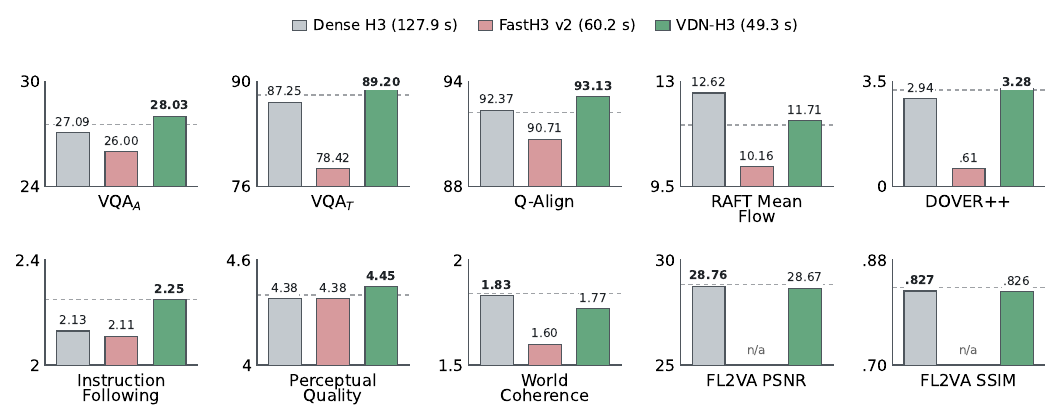}
  \caption{\rev{8-NFE comparison across the same ten metrics. VDN-H3 is higher than FastH3 v2 on all seven shared quality metrics, remains close to Dense H3, and runs 1.2$\times$ faster than FastH3 v2 on one B200. FastH3 v2 was not evaluated on FL2VA; dashed lines mark 50-NFE Dense H3.}}
  \label{fig-quality-8-main}
\end{figure}

\rev{Figure \ref{fig-quality-8-main} compares the 8-NFE samplers on the same ten metrics. VDN-H3 exceeds FastH3~v2 on all seven shared quality metrics, with margins of 10.78 on VQA\textsubscript{T} and 2.67 on DOVER++. It also exceeds 8-NFE Dense H3 on six of nine quality metrics; world coherence and FL2VA endpoint fidelity remain close. Appendix~\hyperref[appendix-quality-steps]{A.6} reports FAST-VQA, FL2VA LPIPS, and the 4-NFE results.}

\rev{Additional paired qualitative samples appear in Figure \ref{fig-gallery}.}

\subsection{Efficiency Results}\label{efficiency-results}

\textbf{Backbone acceleration.} We first compare one complete transformer evaluation at the same sequence length and NFE count. For the 14.3-second, 768p workload, the system-optimized VDN-H3 backbone reduces latency from 35.35 to 11.16 seconds on one H200 (3.2$\times$), and from 16.0 to 6.2 seconds on one B200 (2.6$\times$).

\textbf{Scaling with video length.} Dense Softmax scores every video-token pair, so its video--video workload grows quadratically with the number of latent frames. VDN keeps only a fixed-width local window and the first- and last-frame anchors in Softmax, while VDA carries the remaining long-range context with linear scaling. As the sequence grows from 42 to 102 latent frames, the measured Softmax attention density falls from 42.1\% to 20.0\%. Over the same range, whole-backbone speedup increases from 1.8$\times$ to 3.2$\times$ on H200 and from \rev{1.6$\times$} to 2.6$\times$ on B200. \rev{The duration sweep compares Dense H3 and VDN-H3 on one GPU at the same 50-NFE budget, so these gains isolate backbone acceleration rather than step distillation.}

\textbf{Sampling and distributed inference.} Starting from the optimized backbone, eight-step distillation reduces one-B200 denoising from 307.9 to 49.3 seconds, and eight-GPU head-sharded inference brings the final latency to 6.7 seconds on B200 and 12.5 seconds on H200. Tables \rev{1 and 2} report the complete deployment path and duration scaling on both accelerators.

\begin{table}[t]
\captionsetup{labelfont={sf,bf,color=revisionred}}
\caption{B200 efficiency. Gain is measured against the preceding row; speedup is cumulative against single-GPU Dense H3. \rev{The one-GPU duration sweep compares Dense H3 and VDN-H3 at the same 50-NFE budget, excluding few-step distillation and distributed inference.}}
\label{tab:b200-efficiency}
\centering
\begin{minipage}[t]{0.590\textwidth}
\centering\scriptsize
\setlength{\tabcolsep}{1.5pt}\renewcommand{\arraystretch}{1.18}
\begin{tabular*}{\linewidth}{@{\extracolsep{\fill}}lcccccc@{}}
\toprule
\textbf{Configuration} & \textbf{\#GPU} & \textbf{NFE} & \textbf{s/NFE} & \shortstack{\textbf{End-to-end}\\\textbf{latency}} & \textbf{Gain} & \textbf{Speedup} \\
\midrule
Dense H3 & 1 & 50 & 15.99 s & 799.6 s & -- & $\phantom{00}1.0\times$ \\
VDN-H3 & 1 & 50 & 6.16 s & 307.9 s & $2.6\times$ & $\phantom{00}2.6\times$ \\
$+$ few-step & 1 & 8 & 6.16 s & 49.3 s & $6.3\times$ & $\phantom{0}16.2\times$ \\
$+$ distributed & 8 & 8 & 0.85 s & 6.70 s & $7.4\times$ & $119.3\times$ \\
\bottomrule
\end{tabular*}
\end{minipage}\hfill
\begin{minipage}[t]{0.390\textwidth}
\centering\scriptsize
\setlength{\tabcolsep}{1.5pt}\renewcommand{\arraystretch}{1.18}
\begin{tabular}{@{}lcccc@{}}
\toprule
\shortstack{\textbf{Latent}\\\textbf{frames}} & \textbf{42} & \textbf{72} & \textbf{87} & \textbf{102} \\
\midrule
Attention density & 42.08\% & 26.82\% & 22.85\% & 19.98\% \\
Attention speedup & $1.7\times$ & $2.3\times$ & $2.7\times$ & $3.0\times$ \\
\rev{VDN-H3 latency} & \rev{118.8 s} & \rev{213.1 s} & \rev{260.0 s} & \rev{307.9 s} \\
\rev{End-to-end speedup} & \rev{$1.6\times$} & \rev{$2.1\times$} & \rev{$2.4\times$} & \rev{$\mathbf{2.6\times}$} \\
\bottomrule
\end{tabular}
\end{minipage}
\end{table}

\begin{table}[t]
\captionsetup{labelfont={sf,bf,color=revisionred}}
\caption{H200 efficiency. Gain is measured against the preceding row; speedup is cumulative against single-GPU Dense H3. \rev{The one-GPU duration sweep compares Dense H3 and VDN-H3 at the same 50-NFE budget, excluding few-step distillation and distributed inference.}}
\label{tab:h200-efficiency}
\centering
\begin{minipage}[t]{0.590\textwidth}
\centering\scriptsize
\setlength{\tabcolsep}{1.5pt}\renewcommand{\arraystretch}{1.18}
\begin{tabular*}{\linewidth}{@{\extracolsep{\fill}}lcccccc@{}}
\toprule
\textbf{Configuration} & \textbf{\#GPU} & \textbf{NFE} & \textbf{s/NFE} & \shortstack{\textbf{End-to-end}\\\textbf{latency}} & \textbf{Gain} & \textbf{Speedup} \\
\midrule
Dense H3 & 1 & 50 & 35.35 s & 1767.3 s & -- & $\phantom{00}1.0\times$ \\
VDN-H3 & 1 & 50 & 11.16 s & 557.9 s & $3.2\times$ & $\phantom{00}3.2\times$ \\
$+$ few-step & 1 & 8 & 11.16 s & 89.3 s & $6.3\times$ & $\phantom{0}19.8\times$ \\
$+$ distributed & 8 & 8 & 1.56 s & 12.5 s & $7.1\times$ & $141.6\times$ \\
\bottomrule
\end{tabular*}
\end{minipage}\hfill
\begin{minipage}[t]{0.390\textwidth}
\centering\scriptsize
\setlength{\tabcolsep}{1.5pt}\renewcommand{\arraystretch}{1.18}
\begin{tabular}{@{}lcccc@{}}
\toprule
\shortstack{\textbf{Latent}\\\textbf{frames}} & \textbf{42} & \textbf{72} & \textbf{87} & \textbf{102} \\
\midrule
Attention density & 42.08\% & 26.82\% & 22.85\% & 19.98\% \\
Attention speedup & $2.0\times$ & $3.0\times$ & $3.5\times$ & $4.0\times$ \\
\rev{VDN-H3 latency} & \rev{218.1 s} & \rev{388.8 s} & \rev{476.9 s} & \rev{557.9 s} \\
\rev{End-to-end speedup} & \rev{$1.8\times$} & \rev{$2.5\times$} & \rev{$2.8\times$} & \rev{$\mathbf{3.2\times}$} \\
\bottomrule
\end{tabular}
\end{minipage}
\end{table}

\subsection{Ablation Studies}\label{ablation-studies}

\paragraph{\rev{Architecture ablations}}\label{architecture-ablation}

\rev{Table \ref{tab-architecture-ablation} tests the frame update, boundary anchors, and long-range Linear branch separately. We first replace VDA's joint solve with a frozen-state additive update that uses an extra {$1/\sqrt U$} key scaling factor. VDA scores higher on five of eight quality metrics, including gains of 0.42 on VQA\textsubscript{A}, 0.69 on VQA\textsubscript{T}, 0.15 on FAST-VQA, and 0.07 on FIRM world coherence. The additive update scores higher by 0.19 on Q-Align and 0.14 on DOVER++, while FIRM perceptual quality ties. VDA thus avoids the frame-size-dependent scaling factor while retaining competitive quality and scoring higher on most metrics; its inherited-state transition is also non-expansive. Removing the first- and last-frame anchors while keeping VDA and the Linear branch reduces seven of eight quality metrics. The largest changes are {$-0.75$} in VQA\textsubscript{A} and {$-1.35$} in VQA\textsubscript{T}; Q-Align is the only metric that increases, by 0.24. This supports using the endpoints as explicit global references in addition to the local Softmax window. Finally, comparing the no-anchor VDN variant with Window Softmax only isolates the Linear branch because both omit the anchors. Removing the branch lowers six of eight quality metrics, including VQA\textsubscript{A} by 1.57, VQA\textsubscript{T} by 1.75, and FAST-VQA by 1.77, showing that local Softmax alone does not recover distant context.}

\begin{table}[htbp]
\color{revisionred}
\captionsetup{font={small,color=revisionred},labelfont={sf,bf,color=revisionred}}
\caption{Architecture ablations on video quality and motion.}
\label{tab-architecture-ablation}
\centering\scriptsize
\setlength{\tabcolsep}{2pt}\renewcommand{\arraystretch}{1.16}
\begin{tabular*}{\linewidth}{@{\extracolsep{\fill}}>{\raggedright\arraybackslash}p{2.3cm}*{9}{r}@{}}
\toprule
\textbf{Method} & \textbf{VQA\textsubscript{A}} & \textbf{VQA\textsubscript{T}} & \textbf{Q-Align} & \textbf{FAST-VQA} & \textbf{DOVER++} & \textbf{\shortstack{FIRM\\IF}} & \textbf{\shortstack{FIRM\\PQ}} & \textbf{\shortstack{FIRM\\WC}} & \textbf{Flow} \\
\midrule
VDN & \textbf{30.56} & \textbf{90.11} & 92.69 & \textbf{83.47} & 3.62 & 2.27 & \textbf{4.47} & \textbf{1.83} & 11.67 \\
$-$ Joint VDA solve & 30.14 & 89.42 & 92.88 & 83.32 & \textbf{3.76} & 2.25 & \textbf{4.47} & 1.76 & 11.29 \\
$-$ Boundary anchors & 29.81 & 88.76 & \textbf{92.93} & 83.34 & 3.50 & 2.25 & 4.43 & 1.79 & 11.58 \\
Window Softmax only & 28.24 & 87.01 & 92.18 & 81.57 & 3.24 & \textbf{2.30} & 4.38 & 1.82 & 11.99 \\
\bottomrule
\end{tabular*}
\par\vspace{3pt}\raggedright\scriptsize Bold marks the maximum of each quality column. Flow is a motion diagnostic. ``$-$ Joint VDA solve'' uses frozen-state additive writes with extra frame-size key scaling. Window Softmax only omits both anchors and the Linear branch; the no-anchor VDN row is its controlled comparison.
\end{table}

\noindent\begin{minipage}{\linewidth}
\paragraph{\rev{Stage partitioning}}\label{stage-analysis}

\rev{Table \ref{tab-stage-parameter-drift} measures {$\|\theta_s-\theta_{s-1}\|_2/\|\theta_{s-1}\|_2$} within each stage. Despite using only 200 steps, A1 moves every monitored module the most; the 500-step A2 and 2,000-step Stage B make successively smaller changes, for example {$35.6\!\rightarrow\!17.3\!\rightarrow\!14.0\%$} for the write-gate projection and {$1.25\!\rightarrow\!0.82\!\rightarrow\!0.58\%$} for {$W_o$}. Thus A1 supplies the main branch initialization, while the longer end-to-end stages refine composition and co-adaptation.}
\par\end{minipage}

\begin{table}[htbp]
\color{revisionred}
\captionsetup{font={small,color=revisionred},labelfont={sf,bf,color=revisionred}}
\caption{Stage-wise relative $\ell_2$ parameter displacement (\%).}
\label{tab-stage-parameter-drift}
\centering\scriptsize
\setlength{\tabcolsep}{2pt}\renewcommand{\arraystretch}{1.16}
\begin{tabular*}{\linewidth}{@{\extracolsep{\fill}}lcccccccc@{}}
\toprule
\textbf{Stage} & \textbf{Steps} & \textbf{$\beta$} & \textbf{$g_o$ Down} & \textbf{$g_o$ Up} & \textbf{$\alpha$ Down} & \textbf{$\alpha$ Up} & \textbf{$b_o$} & \textbf{$W_o$ (1.93B)} \\
\midrule
A1 & 200 & 35.6 & 32.9 & 24.7 & 20.5 & 18.3 & 15.3 & 1.25 \\
A2 & 500 & 17.3 & 17.6 & 13.9 & 14.1 & 11.3 & 13.7 & 0.82 \\
B & 2,000 & 14.0 & 14.4 & 11.1 & 12.2 & 9.2 & 9.9 & 0.58 \\
\bottomrule
\end{tabular*}
\par\vspace{3pt}\raggedright\scriptsize Columns track the write-gate projection $\beta$, the down/up projections of output-gate MLP $g_o$ and decay MLP $\alpha$, output-gate bias $b_o$, and output projection $W_o$. Each percentage is measured against its own stage-start parameter norm.
\end{table}

\paragraph{\rev{Kernel-level optimization}}\label{kernel-ablation}

\begin{figure}[!htbp]
\centering
\includegraphics[width=\linewidth]{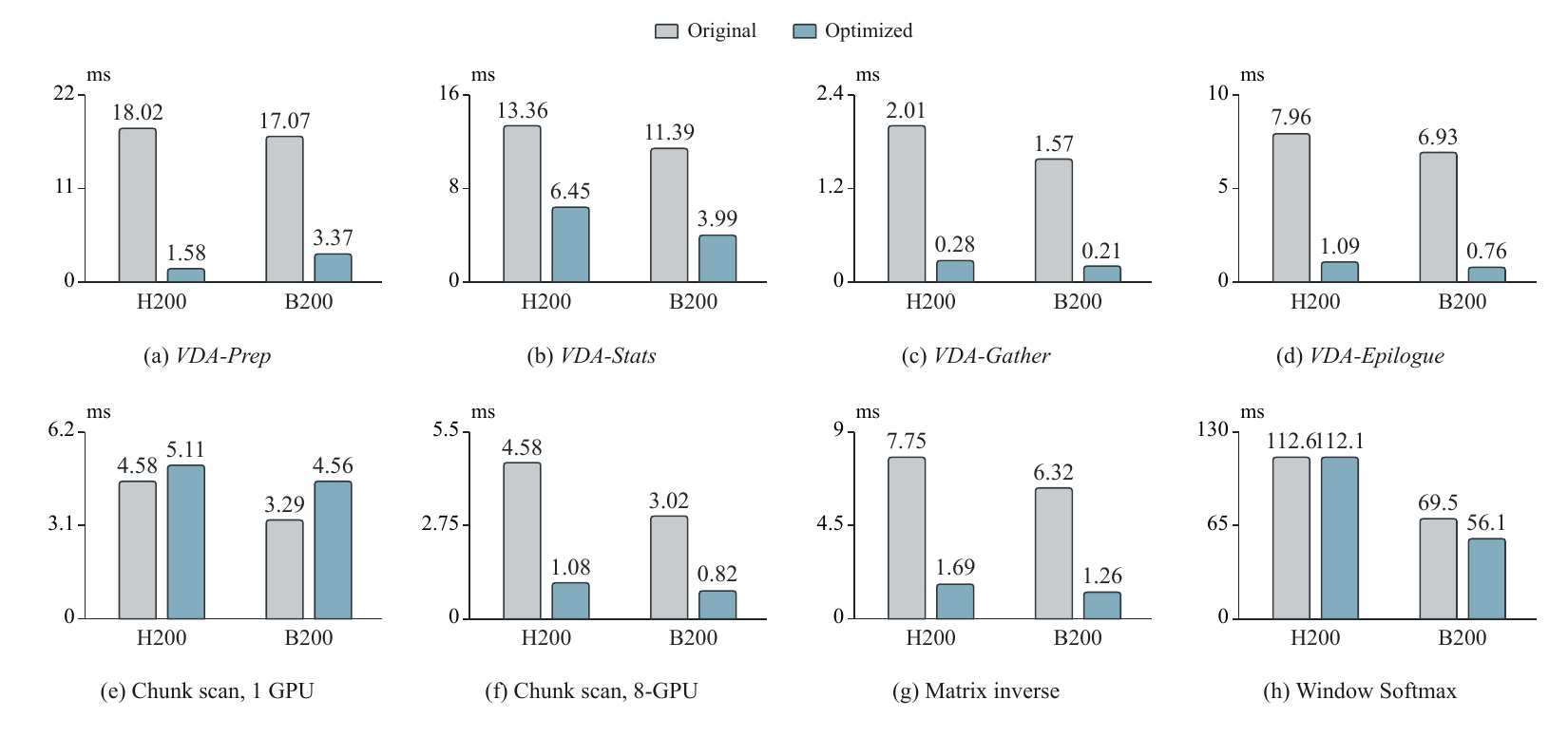}
\caption{Kernel-level optimization ablations on H200 and B200.}
\label{fig-ablation}
\end{figure}

We profile VDN's core backbone operators at 102 latent frames. Figure \ref{fig-ablation} compares their direct and optimized implementations on H200 and B200, separating operator-level gains from few-step distillation and distributed inference.

\textbf{Fused VDA kernels.} \emph{VDA-Prep} reduces latency from 18.0 to 1.6 ms on H200 and from 17.1 to 3.4 ms on B200. \emph{VDA-Stats} provides 2.1$\times$ and 2.9$\times$ speedups, \emph{VDA-Gather} provides 7.2$\times$ and 7.5$\times$, and \emph{VDA-Epilogue} provides 7.3$\times$ and 9.1$\times$ on H200 and B200, respectively.

\noindent\begin{minipage}{\linewidth}
\textbf{Chunk-wise scans.} Composing frame transitions adds overhead with all 56 heads on one GPU, but becomes effective after head sharding. With seven heads per rank in 8-GPU inference, scan latency falls from 4.6 to 1.1 ms on H200 and from 3.0 to 0.8 ms on B200.
\par\end{minipage}

\textbf{Small-matrix inverse.} Replacing the multi-kernel Cholesky path with the fused inverse reduces latency from 7.8 to 1.7 ms on H200 and from 6.3 to 1.3 ms on B200, a 4.6--5.0$\times$ speedup.

\textbf{Other optimizations.} Figure \ref{fig-ablation} additionally isolates Window Softmax. It improves from 69.5 to 56.1 ms on B200, while H200 remains nearly unchanged at 112.6 versus 112.1 ms.

\subsection{Gallery}\label{gallery}

\begin{figure*}[!t]
\centering
\includegraphics[width=\textwidth,height=0.82\textheight,keepaspectratio]{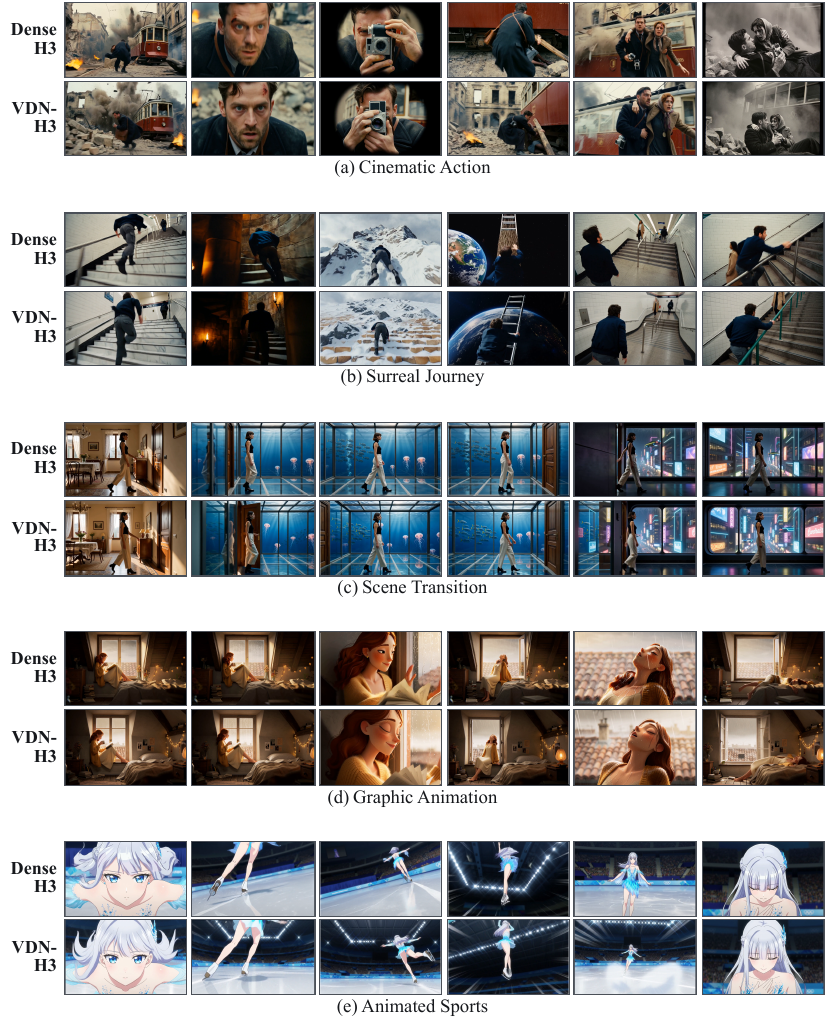}
\caption{Qualitative gallery. Six synchronized frames from five prompts, with 50-step Dense H3 above eight-step VDN-H3 in each pair.}
\label{fig-gallery}
\end{figure*}

In Figure \ref{fig-gallery}, we present diverse examples comparing 50-step Dense H3 and eight-step VDN-H3 across visual styles, scene transitions, and motion patterns. The results demonstrate that VDN-H3 achieves visual quality on par with Dense H3 despite using substantially fewer denoising steps.

\protect\phantomsection\label{related}
\section{Related Work}\label{related-work}

\textbf{Efficient video attention algorithms.} Recent video diffusion transformers process increasingly long spatiotemporal sequences \citep{minimax2026h3,yang2025cogvideox,kong2024hunyuanvideo,wan2025,hacohen2024ltxvideo,polyak2024moviegen,jin2024pyramidflow}. Attention acceleration spans optimized exact kernels \citep{dao2022flashattention,dao2023flashattention2,shah2024flashattention3} and low-precision kernels, including SageAttention \citep{zhang2024sageattention}, SageAttention2 \citep{zhang2024sageattention2}, and SageAttention3 \citep{zhang2025sageattention3}. General-purpose sparse methods include SpargeAttn \citep{zhang2025sparge} and SpargeAttention2 \citep{zhang2026sparge2}; video-specific methods include Sparse VideoGen \citep{xi2025sparsevideogen} and Sparse VideoGen2 \citep{yang2025sparsevideogen2}. SLA and SLA2 combine sparse and linear computation \citep{zhang2025sla,zhang2026sla2}. Quantized diffusion methods reduce the cost of DiT execution \citep{zhao2024viditq,li2024svdquant,li2026deltaquant,xue2026fourtune}, while Quant VideoGen targets the recurrent cache of autoregressive video models \citep{xi2026quantvideogen}.

\textbf{Linear attention and recurrent memory.} Linear Transformers and Performers introduced influential linear-complexity alternatives to dense attention \citep{katharopoulos2020linear,choromanski2021performer}, while Mamba and Mamba-2 connected selective state-space models with efficient sequence processing \citep{gu2023mamba,dao2024mamba2}. Gated Linear Attention \citep{yang2024gla}, DeltaNet \citep{yang2024deltanet}, Gated Delta Networks \citep{yang2024gated}, Kimi Linear \citep{kimiteam2025kimi}, and Gated DeltaNet-2 \citep{hatamizadeh2026gdn2} support data-dependent memory updates. In vision, VideoMamba applies state-space modeling to long-video understanding, and M4V develops a multimodal Mamba backbone for text-to-video generation \citep{li2024videomamba,huang2026m4v}. LoGeR similarly combines sliding-window attention with parametric long-term memory to preserve local geometry and global consistency over long video sequences \citep{zhang2026loger}. Linear and hybrid diffusion backbones include SANA \citep{xie2024sana}, SANA-Video \citep{chen2025sanavideo}, SANA-Video 2.0 \citep{chen2026sanavideo2}, and SANA-WM \citep{zhu2026sanawm}. We also acknowledge Reflections on Video DeltaNet as concurrent analysis of frame-level delta updates \citep{zhu2026reflections}.

\textbf{Few-step distillation and distributed inference.} Progressive distillation, consistency models, latent consistency models, and adversarial distillation reduce the number of denoising evaluations \citep{salimans2022progressive,song2023consistency,luo2023lcm,sauer2023add}. Distribution Matching Distillation \citep{yin2024dmd} and DMD2 \citep{yin2024dmd2} provide closely related objectives; AnimateLCM and T2V-Turbo extend few-step training to video \citep{wang2024animatelcm,li2024t2vturbo}. Multi-GPU inference is complementary: Ulysses and USP shard long attention sequences \citep{jacobs2023ulysses,fang2024usp}, DistriFusion and PipeFusion exploit patch and pipeline parallelism across diffusion steps \citep{li2024distrifusion,fang2024pipefusion}, and xDiT composes multiple parallel strategies in one inference engine \citep{fang2024xdit}. StreamDiffusionV2 further targets low-latency distributed video streaming \citep{feng2026streamdiffusionv2}.

\section{Conclusion}\label{conclusion}

Video DeltaNet accelerates video generation while largely preserving Dense H3 quality. On the 14.3-second, 768p workload, its optimized backbone is 2.6$\times$ faster on one B200 and 3.2$\times$ faster on one H200. With eight-step distillation and eight-GPU inference, DiT denoising takes 6.70 seconds on B200s and 12.5 seconds on H200s. Eight-step VDN-H3 remains close to 50-step Dense H3 across the reported quality evaluations.

\protect\phantomsection\label{appendix}

\bibliographystyle{iclr2027_conference}
\bibliography{references}

\clearpage
\appendix
\section{Derivation and Implementation Details}\label{derivation-and-implementation-details}

\subsection{Proof of Proposition 1}\label{appendix-proof}

Fix the prepared features and gates of frame $t$. Substituting $\bar S_t=S_{t-1}\Diag(\alpha_t)$ into Equation \eqref{eq-solution} separates the inherited state from the new write:

\stepcounter{equation}\begin{equation}\label{eq-transition-comparison}\tag{A.1}
\begin{aligned}
      S_t&=S_{t-1}M_t+J_t,\\
      M_t&=\Diag(\alpha_t)(I+A_t)^{-1},
      \qquad J_t=B_t(I+A_t)^{-1}.
      \end{aligned}
\end{equation}

For two entering states with the same frame inputs, $\Delta S_t=\Delta S_{t-1}M_t$. Since $\beta_{t,u}\ge0$, the Gram matrix is positive semidefinite: $x^\top A_t x=\sum_u\beta_{t,u}(k_{t,u}^\top x)^2\ge0$ for every $x$. Writing $A_t=Q\Diag(\lambda_i)Q^\top$, with $\lambda_i\ge0$, shows that $(I+A_t)^{-1}$ has eigenvalues $1/(1+\lambda_i)\in(0,1]$. Together with $0\le\alpha_{t,j}\le1$, this gives

\stepcounter{equation}\begin{equation}\label{eq-bound}\tag{A.2}
\begin{aligned}
      \|\Delta S_t\|_F
      &\le\|\Delta S_{t-1}\|_F\|M_t\|_2,\\
      \|M_t\|_2
      &\le\|\Diag(\alpha_t)\|_2\|(I+A_t)^{-1}\|_2\le1.
      \end{aligned}
\end{equation}

This proves Proposition 1 without requiring the decay and Gram matrix to commute. The bound concerns inherited state at fixed frame inputs, not new writes or the full network.

\subsection{Frame-size scaling and correlation awareness}\label{frame-size-scaling-and-correlation-awareness}

The difference between independent and joint frame writes can be seen by expanding Equation \eqref{eq-batch-delta}:

\[
S_t^{\mathrm{batch}}=\bar S_t+\sum_{u=1}^{U}\beta_{t,u}(v_{t,u}-\bar S_tk_{t,u})k_{t,u}^{\top}.
\]

All corrections use the same pre-update state, so a patch's residual ignores the other current-frame patches.

By comparison, the first-order condition for VDA can be rearranged as

\[
S_t=\bar S_t+\sum_{u=1}^{U}\beta_{t,u}(v_{t,u}-S_tk_{t,u})k_{t,u}^{\top}.
\]

Here every residual depends on the shared $S_t$. Overlapping keys accumulate in $A_t$, so Equation \eqref{eq-solution} adjusts their joint update.

With $\alpha_t=\mathbf{1}$, the additive inherited-state factor is $I-A_t$. Its eigenvalues $1-\lambda_i(A_t)$ leave $[-1,1]$ if any $\lambda_i(A_t)>2$. For unit keys and sigmoid write gates, $\lambda_{\max}(A_t)\le\operatorname{tr}(A_t)=\sum_u\beta_{t,u}\le U$, and aligned keys can approach this bound. Thus per-token normalization alone does not ensure stability.

\rev{SANA-WM scales each unit key by an additional $1/\sqrt U$ \citep{zhu2026sanawm}. This replaces $A_t$ by $A_t/U$ and $B_t$ by $B_t/\sqrt U$, so the additive inherited-state factor becomes $I-A_t/U$ and is non-expansive under the unit-key, sigmoid-write-gate assumptions above. VDA instead uses $(I+A_t)^{-1}$, whose inherited-state transition is non-expansive for fixed prepared features and gates without frame-size scaling.}

The inverse also responds to the observed key correlations. If all $U$ positions repeat a unit key $k$ with a common gate $\beta$, writing $s=\bar S_tk$ and $\bar v=U^{-1}\sum_u v_u$ gives

\stepcounter{equation}\begin{equation}\label{eq-repeat}\tag{A.3}
S_tk=\frac{1}{1+U\beta}s+
      \frac{U\beta}{1+U\beta}\bar v.
\end{equation}

Repeated keys thus receive a saturating $U\beta/(1+U\beta)$ weight, whereas orthogonal keys each receive $\beta/(1+\beta)$. SANA-WM's scaled-key readout instead assigns $\beta$ to repeated-key consensus and $\beta/U$ to each orthogonal direction. VDA adapts to key geometry rather than applying the same $U$-based scale.

\subsection{Boundary gather and decay bridge}\label{boundary-gather-and-decay-bridge}

After excluding anchor indices 0 and $F-1$, let $P_j$ be the forward state through interior frame $j$ and $R_j$ the reverse state from the last interior frame down to $j$. Their virtual initial states are $P_0=R_{F-1}=S_T/2$. For query $t$, let $[\ell_t,h_t]$ be its local window clipped to interior frame indices. The state read by its prepared queries is

\stepcounter{equation}\begin{equation}\label{eq-bridge}\tag{A.4}
\begin{aligned}
      \widetilde S_t
      =&\ P_{\ell_t-1}\Diag\!\left(\prod_{r=\ell_t}^{t}\alpha_r\right)\\
       &+R_{h_t+1}\Diag\!\left(\prod_{r=t}^{h_t}\alpha_r\right),\\
      o_{t,u}^L=&\ \widetilde S_tq_{t,u}.
      \end{aligned}
\end{equation}

The products are channel-wise. Missing distant video on one side selects the corresponding initial text state rather than a fabricated frame state. The bridge applies decay without local frame writes. If a window covers the complete clip, the implementation bypasses the linear pathway and uses the dense attention path. Clips consisting only of anchor frames also receive no linear video output.

\protect\phantomsection\label{algorithm-vda}
Algorithm 1. Bidirectional VDA inside one hybrid attention block

\begin{enumerate}
\tightlist
\item
  Compute shared QKV and retained multimodal Softmax with chunk and anchor masks.
\item
  Remove anchor video rows from the linear inputs; prepare Q/K/V features and frame gates.
\item
  Compute $A_t,B_t,\alpha_t$ for all interior frames. Form the text state from the prompt.
\item
  Perform a batched factorization of $I+A_t$; construct $C_t,M_t,J_t$ for all frames and heads.
\item
  Scan $S\leftarrow SM_t+J_t$ in both temporal directions, each initialized with $S_T/2$.
\item
  Gather outside-window states and apply the decay bridge in Equation \eqref{eq-bridge}.
\item
  Read each query; apply RMSNorm and the linear gate; zero anchor rows.
\item
  Gate and project Softmax; add the projected linear output to video rows.
\end{enumerate}

\subsection{Architecture adaptation hyperparameters}\label{appendix-training-hparams}

We use AdamW with linear warmup followed by cosine decay. Stages A1, A2, and B use a peak learning rate of $10^{-4}$ and decay to $5\times10^{-6}$; Stage D uses $10^{-5}$ for the generator and $2\times10^{-5}$ for the fake-score model, decaying to $10^{-6}$ and $2\times10^{-6}$, respectively. Following Chimera \citep{ge2026chimera}, parameters are divided into learning-rate groups by module scale: large fan-in matrices use the base schedule, while smaller vector parameters use larger multipliers. In Stage B, the Linear branch uses $0.4\times$ the LoRA learning rate. The table below lists the remaining stage-specific settings.

\noindent\begin{minipage}{0.98\linewidth}
\refstepcounter{table}\label{tab:training-hparams}
\textbf{Table \rev{\thetable}: Hyperparameters for architecture adaptation.} The table covers Stages A1, A2, and B; few-step distillation is described in Section~4.2.
\vspace{4pt}
\centering
\setlength{\tabcolsep}{5pt}
\renewcommand{\arraystretch}{1.28}
\resizebox{\linewidth}{!}{%
\begin{tabular}{@{}p{3.0cm}p{4.4cm}p{4.4cm}p{5.1cm}@{}}
\toprule
\textbf{Setting} & \textbf{A1: per-layer} & \textbf{A2: end-to-end} & \textbf{B: LoRA co-adaptation} \\
\midrule
Parameter status & \textbf{Train:} one Linear branch\newline\textbf{Freeze:} backbone and Softmax gate & \textbf{Train:} all Linear branches\newline\textbf{Freeze:} backbone and Softmax gates & \textbf{Train:} branches, gates, and QKVO LoRA\newline\textbf{Freeze:} backbone FFN, embeddings, and head \\
Steps & 200 & 500 & 2,000 \\
Optimizer / schedule & AdamW; linear warmup $\rightarrow$ cosine & AdamW; linear warmup $\rightarrow$ cosine & AdamW; linear warmup $\rightarrow$ cosine \\
$\beta_1,\beta_2$ & 0.9, 0.999 & 0.9, 0.999 & 0.9, 0.999 \\
Peak / minimum LR & $10^{-4}$ / $5\times10^{-6}$ & $10^{-4}$ / $5\times10^{-6}$ & $10^{-4}$ / $5\times10^{-6}$ \\
Warmup steps & 50 & 50 & 200 \\
Weight decay & 0 & 0 & 0 \\
Gradient clip & 0.1 per layer & 1.0 global & 1.0 global \\
LR multipliers & large 1.0 / small 5.0 & large 1.0 / small 5.0 & LoRA 1.0 / branch 0.4 / small 2.0 \\
LoRA rank / scale & -- & -- & 64 / 64 \\
\bottomrule
\end{tabular}}
\end{minipage}

\subsection{Release configuration and reproducibility}\label{release-configuration-and-reproducibility}

The architectural settings used in this report correspond to the released \texttt{vdn\_solve} branch, chunk size 5 with radius 1, anchors in both row and column directions, K/V short convolution, text-state initialization, and the alpha decay bridge. Architecture metadata is created in A1 and inherited from the checkpoint in later stages. This is preferable to silently changing an attention mask or state rule when resuming a run.

The release package will include exact weight hashes, dataset and evaluation-prompt manifests, random seeds, tokenizer and frame-alignment metadata, GPU topology, software versions, timing protocol, and quality-evaluation outputs. These artifacts connect each reported result to a reproducible model and execution configuration.

\subsection{Quality across denoising budgets}\label{appendix-quality-steps}

Figures \ref{fig-quality-budget-50}--\ref{fig-quality-budget-4} compare the same evaluation suite at 50, eight, and four neural function evaluations (NFEs). Each metric uses the same labeled, truncated vertical scale across the three figures, but different metrics have different scales. An asterisk reproduces the reported paired significance flag relative to 50-step Dense H3; it does not establish significance between two non-dense variants. RAFT mean flow is a motion-magnitude diagnostic rather than a quality score, and lower FL2VA LPIPS is better.

\textbf{Fifty steps.} VDN-H3 after Stage B is broadly on par with Dense H3 before few-step distillation. Their aesthetic, technical, learned-quality, and FIRM scores are close, although two endpoint-fidelity measures show small degradations. This comparison isolates the hybrid architecture and its adaptation from any reduction in sampling steps.

\noindent\begin{minipage}{\linewidth}
\refstepcounter{figure}\label{fig-quality-budget-50}
\centering
\includegraphics[width=.97\linewidth]{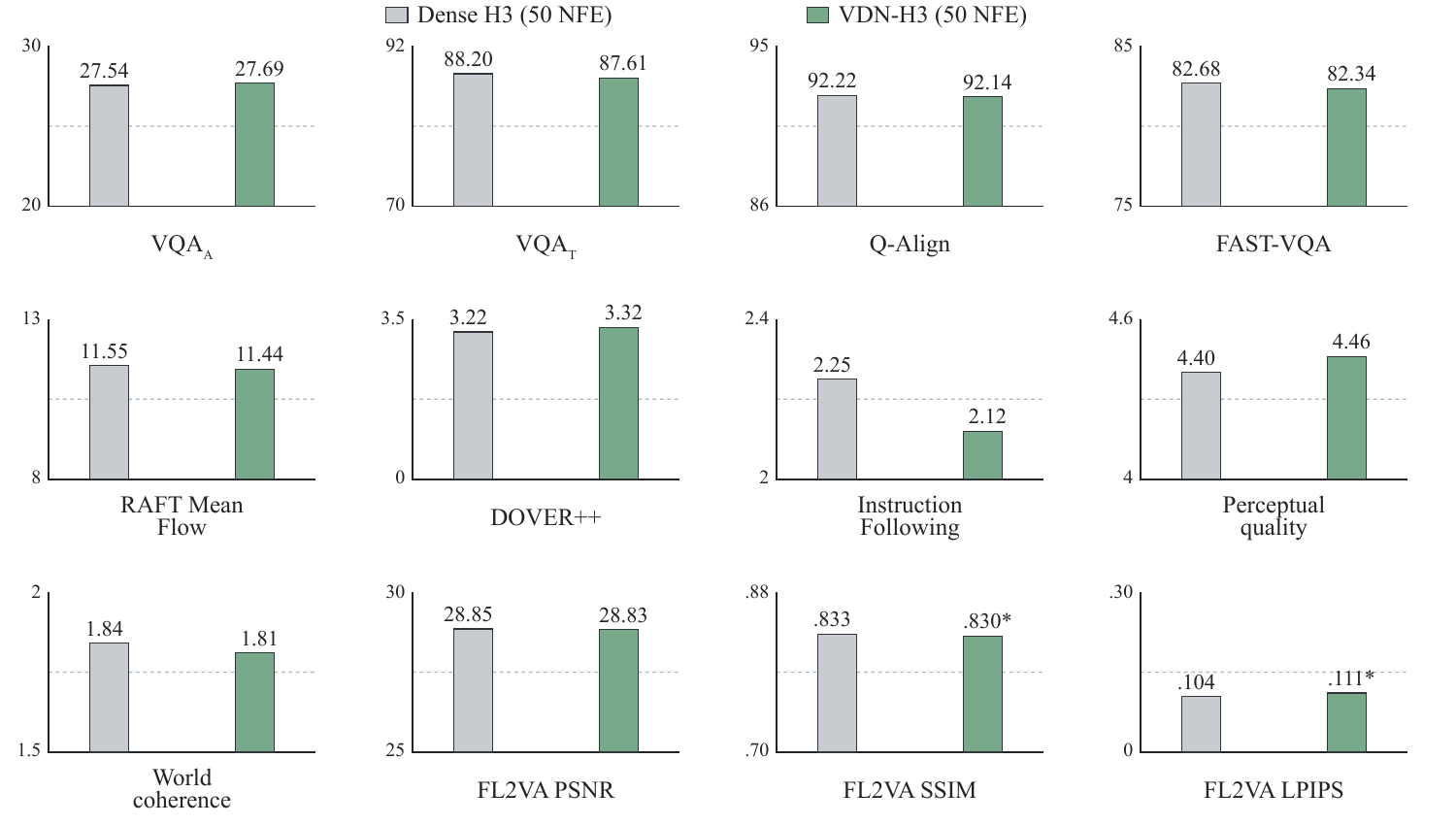}
\par\vspace{2pt}
\small\textbf{Figure \thefigure:} Quality at 50 NFEs before few-step distillation. VDN-H3 is the Stage-B checkpoint.
\end{minipage}

\textbf{Eight steps.} We compare distilled VDN-H3 with Dense H3 plus Larry's eight-step adapter and FastH3 v2 at the same NFE count. VDN-H3 scores highest on all five no-reference quality measures and remains close to the dense-plus-Larry reference on FIRM and FL2VA. Its VQA\textsubscript{T} score is 10.78 points higher than FastH3 v2. FL2VA scores for FastH3 v2 were not available.

\noindent\begin{minipage}{\linewidth}
\refstepcounter{figure}\label{fig-quality-budget-8}
\centering
\includegraphics[width=.97\linewidth]{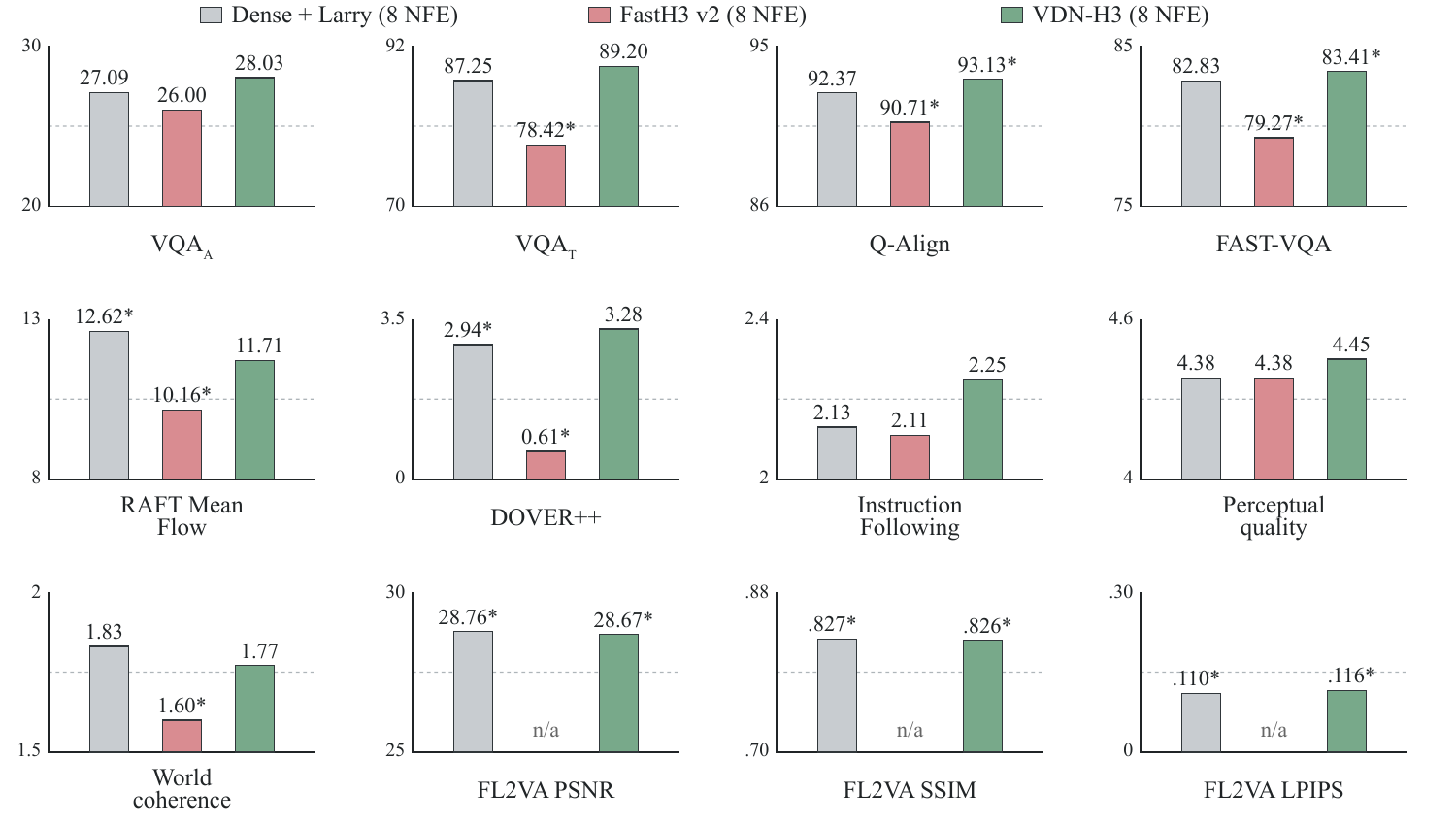}
\par\vspace{2pt}
\small\textbf{Figure \thefigure:} Quality at eight NFEs. FastH3 v2 has no reported FL2VA scores in this comparison.
\end{minipage}

\textbf{Four steps.} VDN-H3 remains compatible with Larry-style few-step adaptation at the lower sampling budget. It scores above both Dense H3 plus Larry and FastH3 v1 on all five no-reference quality measures; against FastH3 v1, its VQA\textsubscript{T} score is higher by 12.64 points, with better FL2VA endpoint fidelity. These are score differences: the supplied significance flags compare each variant with Dense 50, not VDN-H3 directly with FastH3.

\noindent\begin{minipage}{\linewidth}
\refstepcounter{figure}\label{fig-quality-budget-4}
\centering
\includegraphics[width=.97\linewidth]{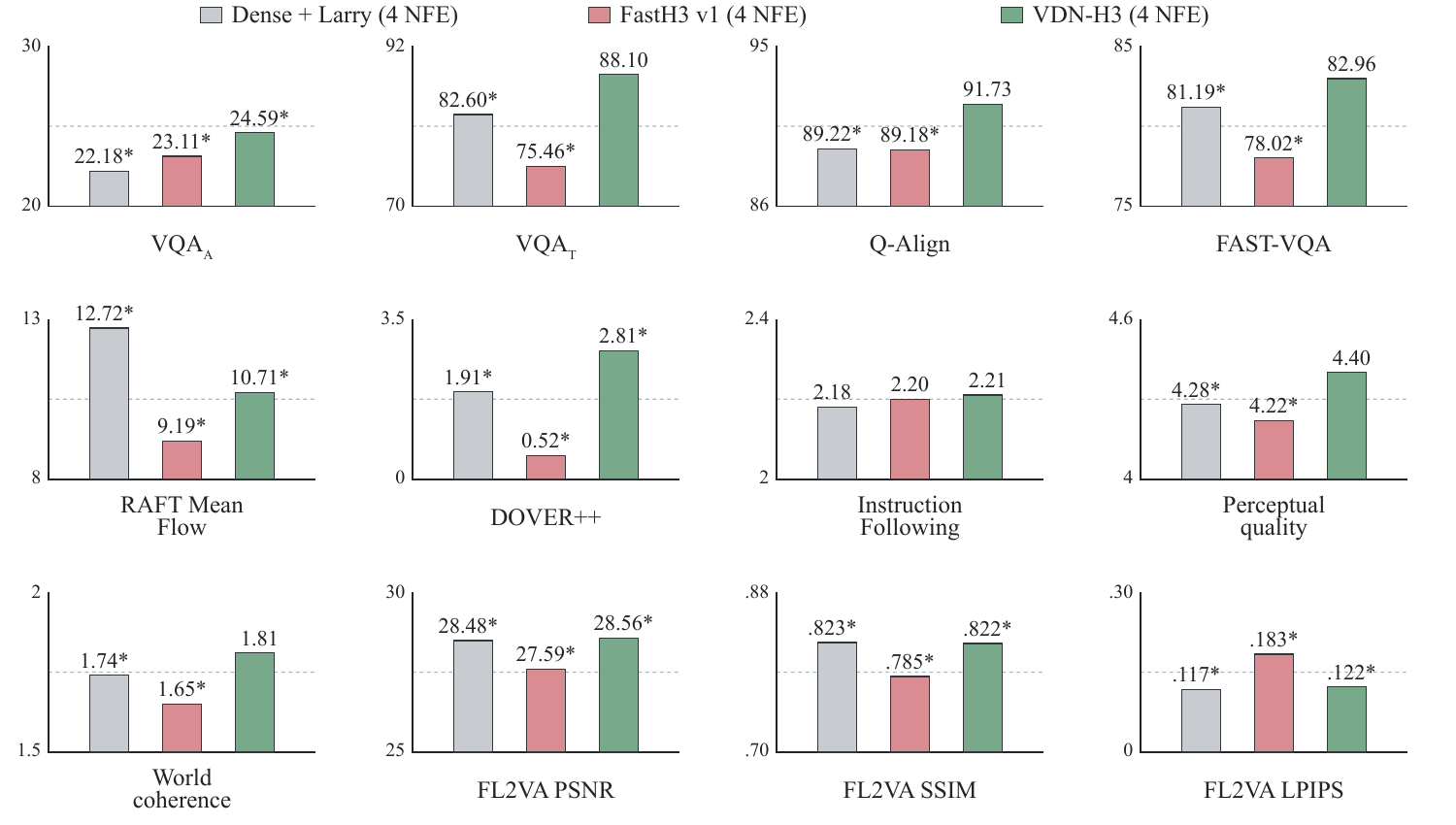}
\par\vspace{2pt}
\small\textbf{Figure \thefigure:} Quality at four NFEs. VDN-H3 retains a clear margin over FastH3 v1 while using the same number of denoising steps.
\end{minipage}

\end{document}